\documentclass[a4paper,twoside]{article}

\usepackage{epsfig}
\usepackage{subcaption}
\usepackage{calc}
\usepackage{amssymb}
\usepackage{amstext}
\usepackage{amsmath}
\usepackage{amsthm}
\usepackage{multicol}
\usepackage{pslatex}
\usepackage{apalike}
\usepackage{algorithm2e}
\usepackage[bottom]{footmisc}
\usepackage{SCITEPRESS}     
\newtheorem{definition}{Definition}

\usepackage{comment}
\usepackage{circledsteps}
\usepackage{tabularx}
\usepackage{array}
\usepackage{listings}

\begin{document}

\title{ Verified Design of Robotic Autonomous Systems using Probabilistic Model Checking}

\author{\authorname{Atef Azaiez\sup{1}\orcidAuthor{0009-0001-5279-686X} and David A. Anisi\sup{1}\orcidAuthor{0000-0003-0870-4259} }
\affiliation{\sup{1}Faculty of Science and Tech., Norwegian University of Life Sciences (NMBU), Ås, Norway}
\email{\{atef.azaiez, david.anisi\}@nmbu.no}
}


\abstract{
Safety and reliability play a crucial role when designing Robotic Autonomous Systems (RAS). Early consideration of hazards, risks and mitigation actions -- already in the concept study phase -- are important steps in building a solid foundations for the subsequent steps in the system engineering life cycle. The complex nature of RAS, as well as the uncertain and dynamic environments the robots operate within, do not merely effect  fault management and operation robustness, but also makes the task of system design concept selection, a hard problem to address. Approaches to tackle the mentioned challenges and their implications on system design, range from ad-hoc concept development and design practices, to systematic, statistical and analytical techniques of Model Based Systems Engineering. In this paper, we propose a methodology to apply a formal method, namely Probabilistic Model Checking (PMC), to enable systematic evaluation and analysis of a given set of system design concepts, ultimately leading to  a set of Verified Designs (VD).  We illustrate the application of  the suggested methodology -- using PRISM as probabilistic model checker -- to a practical RAS concept selection use-case from agriculture robotics. Along the way, we also develop and present a domain-specific Design Evaluation Criteria for agri-RAS.} 

\onecolumn \maketitle \normalsize \setcounter{footnote}{0} \vfill


\section{\uppercase{Introduction}}
\label{sec:introduction}
 
Robotic Autonomous Systems (RAS) may be defined 
as a physical entity, fitted to a moving platform to allow navigating and interacting with the environment, but also equipped with decision-making capability to adapt and independently take decisions to fulfill designated tasks.
RAS are complex systems that require multidisciplinary efforts to design, build, deploy and operate.
 
Several challenges are associated with the design of RAS, namely the uncertain nature of the operational environment, which makes meeting the safety and reliability requirements quite hard to achieve. Another aspect is the technical decision and design choices that influence risk management and mitigation. Some of these choices are typically competing, making some of the safety and reliability properties subject to trade-offs. 
In Addition, a multitude of stakeholders are required to consult with various regulatory and technical sources to be able to specify system requirements.

Some of the current design practices of RAS  are based on ad-hoc, or legacy, builds to reduce development time and cut cost by relying on Commercial Off-The-Shelf (COTS) modules fitted to mobile platforms. This engineering practice can also be noticed in automotive industry and autonomous vehicle development where vehicles get fitted with various perception packages and sensor suites  (e.g. multi camera cluster, lidar, radar to enable the desired autonomy level while fulfilling the required safety levels.
Due to the rapid development of sensor technology and increasing processing power of onboard computer unit, some design changes occur quite late during system- development and implementation or even during operation if certain performance metrics turn out to be below expectations~\cite{Sensors_for_AVs_IEEE_Spectrum}.
 
During the concept selection phase, there exist a panel of frameworks, tools and techniques to facilitate the process of exploring the design space and assessing the suitability of design variants~\cite{Ehlers2025}. A Multi-Criteria Decision Making (MCDM) process is often used to help synthesize the best design concept. MCDM uses tools like Weighted Scoring Models, Analytic Hierarchy Process~\cite{ulker2015} and Pugh Matrix~\cite{vongvit2019}. These tools provide a structured framework to quantify qualitative metrics and allow decision makers to rank and classify potential concepts.

Despite the benefits these MCDM tools and techniques provide to the designers' decision making process, they do not  provide any formal guarantees ensuring that all requirements are satisfied. Furthermore, when taking into account the stochastic properties of the system behavior itself or its environment, which is the main source of uncertainty and risk in the context of RAS design, these tools might lack the necessary expressiveness needed to model the requirements or system behavior.

In order to address this issue, we looked into Formal Methods (FM) which is a discipline of computer science representing a collection of mathematically rigorous techniques used for specification, synthesis, and verification of software and hardware systems~\cite{Woodcock2009}. In recent years, this discipline has gained more popularity among communities seeking to benefit from the mathematically sound, rigorous and verifiable approach FM offers in their respective domains. Also within the RAS domain, recent surveys show a growing use of FM methods in general and (Probabilistic) Model Checking in particular, at various stages of RAS system design~\cite{azaiez2025}. One particularly widely used approach, is Model checking. Probabilistic Model Checking (PMC) in particular has been increasingly used as an approach allowing to synthesize "correct-by-construction" control policies for RAS~\cite{kwiatkowska2025}. However, nor FM in general nor PMC are  typically used at the concept design and selection phase.

  In this paper, we propose a novel approach to add a formal verification layer to the design space exploration step during the concept study phase. Our framework start off by generating initial design alternatives, then systematically run PMC on every variant, and finalize the selection, arriving at a Verified Design fulfilling all the considered requirements.

The remainder of the paper is organized as follows. Section~\ref{sec:related_work} provides an exposition of related work. In Section~\ref{sec:methodology}, we present the generic methodology which can be applied to any system presenting similar properties to RAS by combining FM approach, in particular PMC,  
with MCDM methods. In Section~\ref{sec:implementation}, we apply this methodology to a particular use-case  in the context of RAS design with an agricultural multi-arm agricultural robot. Finally, we reflect on the paper and results by providing
concluding remarks and discussions on future work in Section~\ref{sec:conclusion}.


\section{\uppercase{Related Work}}
\label{sec:related_work}
 
Paper~\cite{ceisel2014} illustrates a typical system design workflow for an Unmanned Arial Vehicle (UAV). The process involves application of Model Based Systems Engineering (MBSE) coupled together with MCDM in order to achieve a design candidate compliant with defined system requirements. In contrast to our work, the use of MCDM here-within was limited to \emph{defining affordable requirements}, rather than design concepts. Also, no formal approach and verification was considered. Nevertheless,~\cite{ceisel2014} illustrates the validity of using MCDM techniques when the goal is to select a suitable design variant among a set of alternatives. 

Paper~\cite{phillips2023}, applies model checking principles to formally verify whether a \emph{given} design alternative satisfies given specifications.   
 This work differs for ours, which instead considers both selection \emph{and} verification of a system design among a set of candidates.  In addition, despite offering a valuable improvement to the correctness of the design, it is worth mentioning that the complex nature of RAS especially when taking into account the dynamic and stochastic nature of the environment in which RAS operate, makes modeling certain behavioral aspects of the system quite challenging to capture by only using SysML and subsequently Finite State Machine (FSM). To address this, our work instead utilizes a probabilistic Markovian representation of the system behavior to provide a more realistic modeling and consequently yields to a  more accurate evaluation of design alternatives.

Other contributions have focused on using PMC to \emph{formally synthesize safety controllers} for RAS. The workflow presented in~\cite{gleirscher2020} covers various stages of RAS design although the focus was aimed at risk and mitigation modeling, followed by controller synthesis, rather than the system design verification as in our paper. 
Similarly, as far as focus on designing robotic controllers is considered,~\cite{RoboChart2024} utilizes a state-machine based notation, namely RoboChart, and the dedicated graphic engineering tool (called RoboTool) which enables automatic generation of code and mathematical models from the designed controllers. It is noteworthy however that both~\cite{gleirscher2020} and~\cite{RoboChart2024} consider formal design and synthesis of a safety controller, instead of formal system design, which is the scope of the paper at hand.

\begin{figure}[htbp!]
  \centering
   \includegraphics[width = 7.5cm]{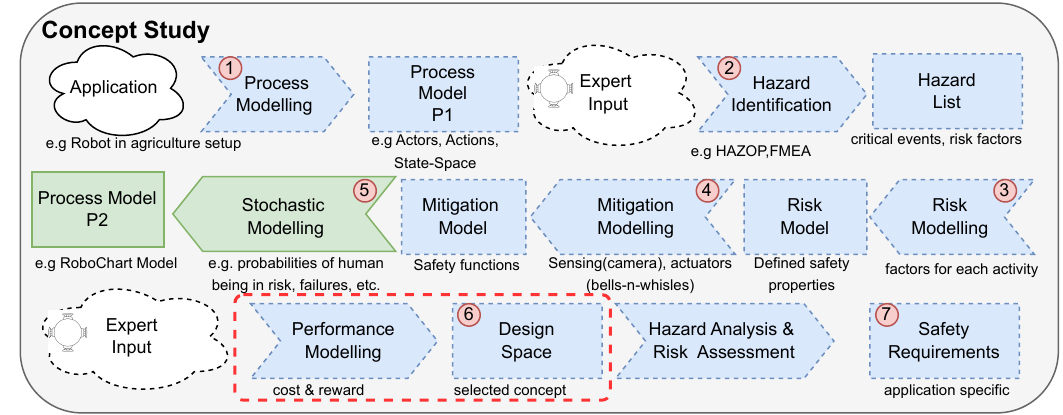} 
  \caption{The concept study is the initial phase of the verification methodology across the engineering life-cycle taken from~\cite{adam2025}. Notice that the performance modeling and Design space generation steps - encircled in red, dashed box -  are non-formal.  }
  \label{fig:meth_scope}
\end{figure}

The work in~\cite{adam2024,adam2025}, present an end-to-end framework for integration of FM and verification steps into   
 the entire RAS development life cycle. The performance modeling and design-space exploration phases of the framework presented in~\cite{adam2025} and highlighted in red dashed box in Figure~\ref{fig:meth_scope}, do not include any formal elements however. 

 Review of related work has thus identified a step of the Concept Study block that is not typically formally verified and we consider that it can be greatly improved by applying FM approaches to make the outcome more robust. In other words, our work is a direct extension of~\cite{adam2025} by incorporating FM into the concept-study phase, arriving at a Verified Design. The proposed approach consists of the Performance Modeling and Design Space Concept Selection steps, marked by dashed, red line in Figure~\ref{fig:meth_scope} and detailed further in the next section.

\section{\uppercase{Methodology}}
\label{sec:methodology}

In this section, we present our developed methodology and demonstrate how to leverage FM and MCDM techniques in the concept study phase to generate \textit{Verified Design}. 
 
Although the methodology is  generic in nature and described as such in this section, we select a specific agricultural RAS use-case to exemplify it in Section~\ref{sec:implementation}. 
 
\begin{figure*}[htbp!]
  \vspace{-0.2cm}
  \centering
   \includegraphics[width=15cm]{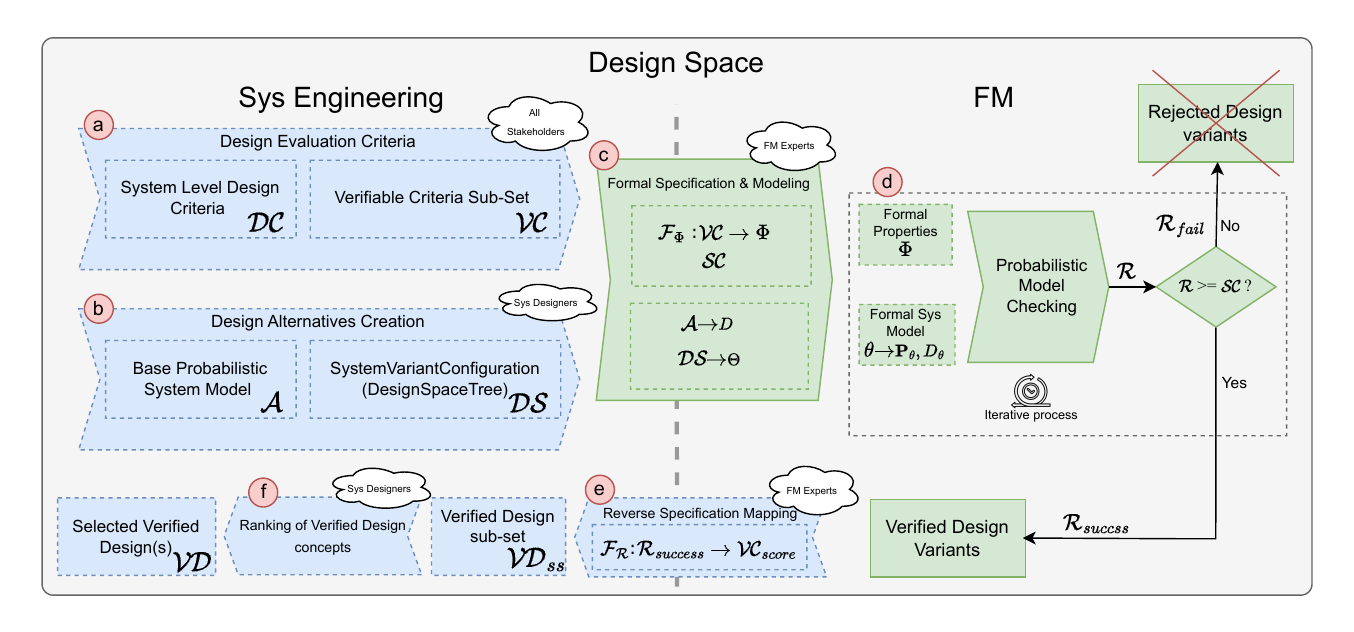}   
  \caption{Methodology workflow: The arrow shapes indicate execution steps, the rectangular shapes indicate artifacts. The blue color marks systems engineering while the green color marks formal methods domain of expertise. } 
  \label{fig:meth_horizontal}
\end{figure*}

Figure~\ref{fig:meth_horizontal} illustrates the main steps involved in the methodology. It is split with a vertical, dashed line into two sides. 
The left hand side represents the non-formal steps and artifacts related to conventional (SOTA) Systems Engineering  practices, whereas the right hand side includes all the FM steps. The methodology begins with \textit{Design evaluation criteria}~\Circled[inner color=blue]{a} and \textit{Design Alternatives Creation}~\Circled[inner color=blue]{b}. We then transform the resulting artifacts into formal language and systematically execute PMC to every design alternative in step~\Circled[inner color=blue]{c}. We further evaluate the results from the PMC step~\Circled[inner color=blue]{d} and eliminate any design alternative that fail the property verification check. In step~\Circled[inner color=blue]{e}, we transform the MC results into scaled values that can be mapped to the specific design criteria and thus enable an effective execution of MCDM step, using existing tools. Finally, by aggregating formally verified and non-formally verified scored criteria, we are in a position to be able to systematically rank the verified design concepts and synthesize a \emph{Verified Design} at step~\Circled[inner color=blue]{f}.

In the remainder of this section, we  provide a more thorough description of each key step. To ease future referencing and understanding, Table~\ref{table:notations} summarizes the  notation adopted in this paper.

\begin{table}[htbp!]
\centering
\caption{Notations and abbreviation description} \label{tab:notations}
\label{table:notations}
\begin{tabular}{|>\centering m{0.07\textwidth}| m{0.35\textwidth}|}
 \hline
      Symbol &   \multicolumn{1}{c|}{Description} \\
 \hline
 \hline
   \( \mathcal{DC} \) & Design Criteria set  \\
\hline
  \( \mathcal{VC} \) & Verifiable Criteria sub-Set  \\
\hline
  \( \mathcal{VC}_{score} \) & Verifiable Criteria Scores  \\
\hline
 \(\mathcal{A}\) & Abstract probabilistic transition system \\
 \hline
 
   \( \mathcal{DS} \) & Design space alternatives  \\
\hline
 \( \Phi\) & Formal property formulae set\\ 
 \hline
 
\( \mathbf{P}_{Res} \)  & Vector of property probability result  \\
 \hline
 \( \mathcal{R} \) & Aggregated probabilistic MC results \\
 \hline
  \( \mathcal{R}_{success} \) & Aggregated \( \mathcal{R} \) at least matched \( \mathcal{SC} \)  \\
 \hline
  \( \mathcal{R}_{fail} \) & Aggregated \( \mathcal{R} \) failed to match \( \mathcal{SC} \) \\
 \hline
\( \Theta \) & Set of all variant configurations \\
 \hline
\( \theta \) & Member of  configuration set, \( \theta \in \Theta   \)  \\
 \hline

\( S \) & Set of states in the DTMC \\
 \hline
 
\( \mathbf{P}_\theta \) & Transition matrix under configuration \( \theta \) \\
 \hline
\( L \) & Labeling function \\
 \hline
\( D \) & Discrete Time Markov Chain (DTMC)  \\
 \hline
 \( D_\theta \) & DTMC instance of design variant \\
 \hline
 \( \mathcal{E} \) & Execution wrapper \\
 \hline
  \( \mathcal{F}_{\Phi} \) & \( \mathcal{VC} \) to \( \Phi \) transformation function \\
 \hline
   \( \mathcal{F}_{\mathcal{R}} \) & \( \mathcal{R} \) to \( \mathcal{VC}_{score} \) transformation function \\
 \hline
   \( \mathcal{VD}_{ss} \) & Verified Design sub-Set  \\
\hline
  \( \mathcal{VD} \) & Verified Design(s)  \\
\hline
\( \mathcal{SC} \) & Success Criteria set \\
\hline
\end{tabular}
\end{table}

\subsection{Design Evaluation criteria~\Circled[inner color=blue]{a}}
The purpose of this step is to create a set of criteria to serve as a ranking basis for the system design variants. Each variant will be allocated a set of scores corresponding to the respective criterion. The results will be later processed by the MCDM process of choice using already available tools at hand. 
Two sub-activities need to be performed during this step. First, we start by creating a system level design criteria set \( \mathcal{DC} \), followed by  selecting only the verifiable sub-set \( \mathcal{VC} \) to be used for model checking.  
\subsubsection*{Design Criteria Specification}
 
Designing RAS typically requires involvement of multiple stakeholders to cover all relevant aspects. Some of the internal stakeholders include design-, development- and safety engineers, project- and product managers and Quality Assurance resources, as well as FM experts. The external stakeholder are mainly end users, suppliers, investors and regulatory bodies. 
As domain experts, stakeholders play a crucial role throughout the design process of RAS.
All stakeholders contribute to the specification of the system level Design Criteria set \( \mathcal{DC} \) that will serve to evaluate various design concepts as indicated in Figure~\ref{fig:meth_horizontal}. Care must be taken when identifying the criteria so that they correspond to the level of abstraction adopted to model the system. In this process, any qualitative criteria have to undergo a quantification process in order to be useful for ranking and comparison of design alternatives. 

\subsubsection*{Design Criteria Selection}
 
This sub-activity is dedicated to identifying design criteria applicable to model checking. 
Hence, only the sub-selection from \( \mathcal{DC} \) of formally verifiable criteria identified as Verifiable Criteria sub-set \( \mathcal{VC} \) is for the next stage. The most important criteria that require rigorous verification are those related to safety and performance of the system. The stakeholders identify the corresponding relevant formal specifications and properties to prepare for the subsequent steps. The rest is kept as is and will be treated further with non-formal methods, not contributing to formal verification of the design model.

\subsection{Design Alternatives Creation~\Circled[inner color=blue]{b}}
In this step, systems designers follow the appropriate abstraction level to create the various design alternatives represented in a formal language and therefore allowing the model checker engine to verify each alternative model against the specifies properties.   
This step is split into two sub-activities as well. First, a base behavioral system model \(\mathcal{A}\) needs to be created in order to establish the system states and the transition logic between these states. Then, depending on the sub-system potential alternatives, a Design Space tree, \( \mathcal{DS} \), is created reflecting all possible combinations.   
 
\subsubsection*{ Probabilistic Base System Modeling}
Starting off from the stochastic process model denoted  P2 in Figure~\ref{fig:meth_scope}, the system is modeled at an appropriate and functional level of abstraction by a state machine described an abstract probabilistic transition system, \(\mathcal{A}\). The process of creating the model P2 is outside the scope of this work, but it is important to emphasize that special care and consideration need to be taken in order to build a balanced model expressive enough to capture system functionality but also manageable in size and complexity to avoid issues related to the nature of model-checking like state explosion. More complex systems can be sub-divided into hierarchical structure to achieve self-contained smaller models. At this stage, the states of the system shall remain unchanged. Any suggested design alternative must only affect the probability values of the transition from one state to the other. Depending on the number and effect of design alternatives, some transition probabilities will remain constant and unaffected by the choice of design alternative, whereas other transition probabilities will need to be set as parameters and will affect the system behavior.

\subsubsection*{System Variant Configuration}
System designers propose technical solutions and create design alternatives to fulfill design requirements. Considerations from Mitigation Modeling  step denoted  ~\Circled[inner color=blue]{6} in Figure~\ref{fig:meth_scope} may also lead to the proposal of additional alternatives for the overall system design. Each sub-system can have an arbitrary number of alternatives. This results in the creation of a “tree” of system variants denoted as \( \mathcal{DS} \). The total number of variants in the simplest form equals to the product of alternatives of each sub-system.%
These alternatives affect the behavior (and the performance) of the system and thus result in different probability of transitioning from one state to the other. As an illustrative example, we can consider two perception sub-system design alternatives of a RAS, where the first alternative consists of a stereo camera system coupled with a machine learning algorithm to identify obstacles and assist with navigation. The second alternative would be complemented with a high resolution 3D lidar system. In this case, with the first alternative, the system will transition the \{DONE\} state  with a higher likelihood compared to the second alternative. This is due to the superior reliability and/or performance a lidar system  may  offer compared to a camera based system operating in a dynamic environment, e.g., with changing lighting conditions. In practice, the specific transition probability values may  be decided by various approaches: Empirical data from reliable sources and test labs, manufacturer documentation and reports, practical benchmarking of each variant and field test results of prior prototypes. A combination of these approaches can also lead to more precise values. In all cases, it is important to note that all these values are subject to future measurement and monitoring during the various field tests as well as during normal operation, so that any substantial deviation can be appropriately addressed.

\subsection{Formal Specification and Modeling~\Circled[inner color=blue]{c}}
This step is crucial as a preparation to the application of model checking to the modeled specifications and system. Stakeholders with FM expertise capture accurate and formal representation of both system and properties on an appropriate level of abstraction. This is quite important, as the model checker will specifically verify all formal properties \( \varphi_i  \in \Phi \) against the system model represented by a Discrete-Time Markov Chain (DTMC)~\cite{baier2008}, denoted \( D \), and any mismatch between the original system requirements  and their corresponding formal specification will lead to inaccurate results. 

\subsubsection*{Design Criteria to Formal Properties Transformation }

We create a transformation function, \( \mathcal{F}_{\Phi} \),  to specify verifiable design criteria, \( \mathcal{VC} \), into Probabilistic Computation Tree Logic (PCTL) formal properties. Some criteria can be expressed in multiple properties or congregated in one property. We also define the Success Criteria set (\( \mathcal{SC} \))  extracting and specifying required property values or ranges so that the model checker can determine if the  properties hold or not.

\subsubsection*{Formal System Model} The base probabilistic system model  specified as an artifact in the previous Design Alternative Creation step~\Circled[inner color=blue]{b}, serves at the base DTMC model, which is further modeled as a parametric DTMC in formal language. The transition probabilities between states are set to constant values specified by System Designers whilst the probabilities subject to different design alternatives are set to be parametric. Care must be taken to make sure that the sum of outbound transitions from any state is equal to 1, \emph{i.e.}
\[ \sum_{s' \in S} P(s,s') = 1  \quad \forall   s \in S. \]
The system variant configuration tree, \( \mathcal{DS} \), is transformed into a number of transition matrices equal to the total number of design space alternatives, \( \Theta \) .

\subsection{Probabilistic Model Checking~\Circled[inner color=blue]{d}}

The Probabilistic Model Checker will run the verification process of the DTMC with all transition matrices \( \mathbf{P}_\theta \) against the formal properties \( \Phi \), one by one in a systematic fashion as described by Algorithm~\ref{alg:mc_alg}. It ultimately returns a result report with all properties verification values.  
Only the subset of the design space that passed the model checking phase will be considered.

\begin{algorithm}[!h]
\caption{Parametric Model checking of Design Space DTMC }
\label{alg:mc_alg}
\KwIn{\( \Phi\), \( \mathcal{DS} \), \( \D \)}
\KwOut{$\mathcal{R}_{sucess} , \mathcal{R}_{fail} $}

Initialize $\mathbf{P}_{\text{default}} \leftarrow$ default transition parameters\;

Generate parameter space $\Theta$ from \( \mathcal{DS} \);\\  
Initialize results $\mathcal{R}, \mathcal{R}_{sucess} , \mathcal{R}_{fail}  \leftarrow \emptyset$\;

\ForEach{$\theta \in \Theta$}{
    Derive $\mathbf{P}_\theta$ from configuration $\theta$;   

    \If{$\exists  s \in S$ such that $\sum_{s' \in S} \mathbf{P}_\theta(s,s') \neq 1$}{
        Log error: violating states\;
        
    \Return
    }
 
        /*sanity check successful*/ \\
        Derive $\D_\theta$ from  $\mathbf{P}_\theta$\;
        Execute PMC:
        \\ $\mathcal{R} \leftarrow \texttt{run} (\D_\theta,  \Phi) $\;  
    
        \If{execution successful}{
            \If { $\mathcal{R} \ge \mathcal{SC} $ }
                {
                 
                 Append $\mathcal{R}$ to $\mathcal{R}_{success}$\;
                }\Else{
                   
                  Append $\mathcal{R}$ to $\mathcal{R}_{fail}$\;
                }
             
        }
        \Else{
            Log execution error\;
             
        }
     
}
\Return { $\mathcal{R}_{sucess}, \mathcal{R}_{fail}$} 
\end{algorithm}

\subsection{Reverse Specification Mapping~\Circled[inner color=blue]{e}}
 
We use a reverse specification transformation function, \( \mathcal{F}_{\mathcal{R}} \), to transform the formal properties results, \( \mathcal{R} \), generated from the probabilistic model checker into score values, \( \mathcal{VC}_{score} \). This step yields a a sub-set of Verified Design variants, \( \mathcal{VD}_{ss} \), with their Verifiable Criteria ready to be ranked and sorted.

\subsection{Ranking of Verified Design concepts~\Circled[inner color=blue]{f}}  
We combine the scores \( \mathcal{VC}_{score}\) of  Verified Design variants sub-set \( \mathcal{VD}_{ss} \) originating from the PMC step together with the the non-formal criteria.
We then apply a MCDM process to the aggregated scored variants in order to be sorted and re-evaluated by the stakeholders by means of tools such as radar graphs. An agreed selection criteria, e.g., surface maximization, can then be used to select the (pareto)optimal design(s).


\section{\uppercase{Agri RAS implementation }}
\label{sec:implementation}

In this section we apply the methodology and algorithm presented Section~\ref{sec:methodology} to a practical use case. The choice of an agricultural RAS application stems from the increasing need for more precise and effective means of crop management and monitoring in the agriculture industry. This context presents some particular challenges that are not as common in other fields especially when combined together. Some of those constraints are manifested  when RAS  are deployed in a dynamic environment such as: 
\begin{itemize}
\item {Changing ambient lighting conditions.}
\item {Uneven terrain and soil composition.}
\item {Plants with rapidly changing physical properties.}
\end{itemize}
 
 The agri-RAS is to perform various inspection and manipulation tasks on various horticulture , such as strawberry and tomato plants. The intended tasks range from runner pruning and improving canopy density, to plant health treatments, e.g., using dedicated UVC-light~\cite{adam2023case}. 

\begin{figure}[hbtp!]
  \vspace{-0.2cm}
  \centering
   \includegraphics[width = 5.5cm]{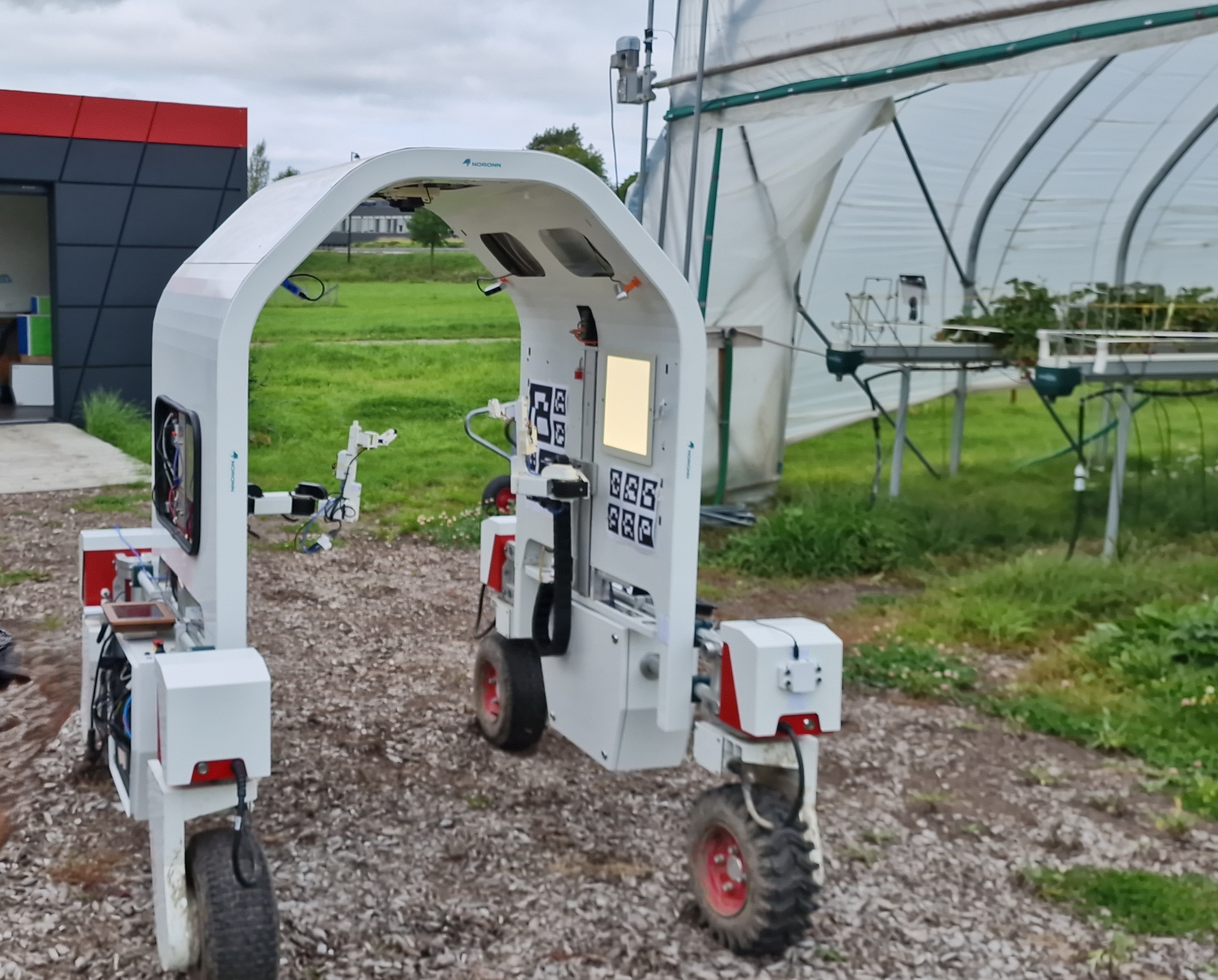}  
  \caption{Thovald robot equipped with two 5DOF arms in a poly-tunnel strawberry test field at NMBU Campus.}  
  \label{fig:sys_presentation}
\end{figure}

The core components of the RAS consist of the following sub-systems:
\begin{itemize}
\item {\textbf{Mobile platform:} A platform enabling locomotion based on the Thorvald multipurpose, omnidirectional and modular agricultural robot.} 
\item {\textbf{Robot-arm(s) and actuators:} One or multiple high-dexterity robotic arms and their respective controllers. These arms are used to perform various inspection and manipulation tasks. To this end, different end-effectors will be fitted on them, depending on the requirements of the task. For example, one arm could be fitted with a combined eye-in-hand camera and cutting device, while a second one could be equipped with a grasping and/or picking tools.}
\item {\textbf{Perception:} A perception system that provides visual scanning of the scene and geometric acquisition. This module is responsible for reconstructing a high-fidelity and accurate 3D model of the robot's work area, including the plants.}
\item {\textbf{Sensor data analysis:} A sensor data processing unit capable of performing identification and classification of objects. }

\end{itemize}
Figure~\ref{fig:sys_presentation} illustrates a prototype of the RAS used for experimentation fitted with robot arms and perception system consisting of four RGBD cameras and lidar scanning devices.
 
In our case, the system can be modeled as a labeled transition system, \( \mathcal{A}\), according to Definition~\ref{def:ts}.

\begin{definition}[Transition System, $(TS)$]
\label{def:ts}
A \emph{labeled transition system}, \( TS \), is a tuple \( (S,Act, \rightarrow, I, AP, L) \) where:
\begin{itemize}
    \item \( S \) is a set of states ("state space"),
    \item \( \mathit{Act} \) is a set of actions,
    \item \( \rightarrow \subseteq S \times S \) is a transition relation,
    \item \( I \subseteq S \) is a set of initial states,
    \item \( AP \) is a set of atomic propositions, and
    \item \( L : S \rightarrow 2^{AP} \) is a labeling function.
\end{itemize}
\end{definition}

\subsection{Design Evaluation Criteria~\Circled[inner color=blue]{a}}

\begin{figure}[!h]
  \vspace{-0.2cm}
  \centering
   \includegraphics[width = 7cm]
   {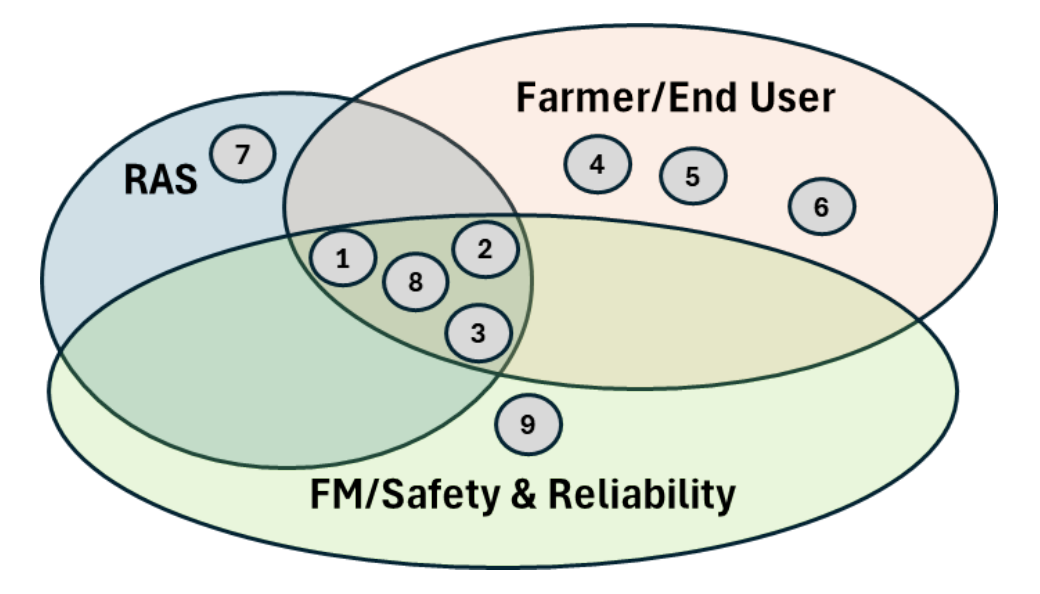}  
  \caption{Design evaluation criteria according to the stakeholders' areas of expertise. The circled numbers correspond to criterion ID described in Table~\ref{table:design_criteria}. }
  \label{fig:venn_criteria}
\end{figure}

All stakeholders involved in the RAS design, depending on their area of expertise are to establish the set of criteria against which the design variants will evaluated. This is a collective effort between all parties involved. Some criteria may have more than a single effect on the design. Figure~\ref{fig:venn_criteria} illustrates the main categories of stakeholders involved in the design of agri-RAS. We broadly can designate each category as follows: 
\begin{itemize}
\item RAS experts: Robot system designers and engineers responsible for the establishment of the technical solutions of all RAS subsystems.  
\item FM/Safety and Reliability experts: Responsible for safety management procedures and implementation of these procedures. This includes interpreting the regulatory safety documents and standards, the safety audits and reviews, as well as for the formal specification and modeling, executing model checking, but also executing the reverse specification mapping to interpret the results coming out of the model checking operation (steps \Circled[inner color=blue]{c}, \Circled[inner color=blue]{d} and ~\Circled[inner color=blue]{e} in Figure~\ref{fig:meth_horizontal}).   
\item Farmers/End Users: Originators of high level system requirements and validator of added value.   
\end{itemize}

\subsubsection*{Design Criteria Specification}

In this paper, we propose a design criteria set that can be adopted as a generic template of agri-RAS design. Our design criteria set  \( \mathcal{DC} \{C_1,C_2...C_9\}  \) is described in Table~\ref{table:design_criteria} and will be used when ranking the design variants using a MCDM process. We will later use a radar graph so that each criterion will represent one axis of the graph. 
\begin{table}[htbp!]
\caption{Design criteria: All criteria that can are formally verifiable at design phase, are marked with a cross (x) in the last column.}
\centering
\label{table:design_criteria}
 
 \begin{tabular}{|>\centering m{0.02\textwidth}| m{0.35\textwidth}|m{0.015\textwidth} |}  
  \hline
  \textbf{\(\mathcal{DC}\)}   & \multicolumn{1}{c|}{Criteria}          & FV  \\  
  \hline
  \hline
  \(C_1\) & Reliability / Safety Property     & x \\
  \hline
  \(C_2\) & Collision Avoidance               & x \\
    \hline
  \(C_3\) & Accuracy, e.g., of perception system  & x \\
    \hline
  \(C_4\) & Performance, e.g., mission success rate & x \\
    \hline
  \(C_5\) & Cost/ROI                          &   \\
    \hline
  \(C_6\) & Mass, Dimensions (physical properties) &  \\
    \hline
  \(C_7\) & Level of Autonomy  &  \\
    \hline
  \(C_8\) & TRL &  \\
    \hline
  \(C_9\) & Computational Complexity  &  \\
  \hline
\end{tabular}
\end{table}

\subsubsection*{Design Criteria Selection}
As explained in the methodology Section~\ref{sec:methodology}, not all criteria can be formally verified. In Table~\ref{table:design_criteria}, the formally verifiable criteria are the ones identified with a cross (x) in the FV column. These form the verifiable sub-set \( \mathcal{VC} = \{C_1,C_2,C_3,C_4 \} \) that will be specified into formal properties in step ~\Circled[inner color=blue]{c}.

\subsection{Design Alternatives Creation~\Circled[inner color=blue]{b}}
We proceed by modeling the RAS such that its behavior is represented by a probabilistic transition model, \(\mathcal{A}\). For the sake of simplicity, we consider that the states represent a high level abstraction of the system while under the influence of one of its sub-systems at any given time step.  
\subsubsection*{ Probabilistic Base System Modeling}
 
\begin{figure*}[!hbtp]
  \vspace{-0.2cm}
  \centering
   \includegraphics[width = 12cm]
  {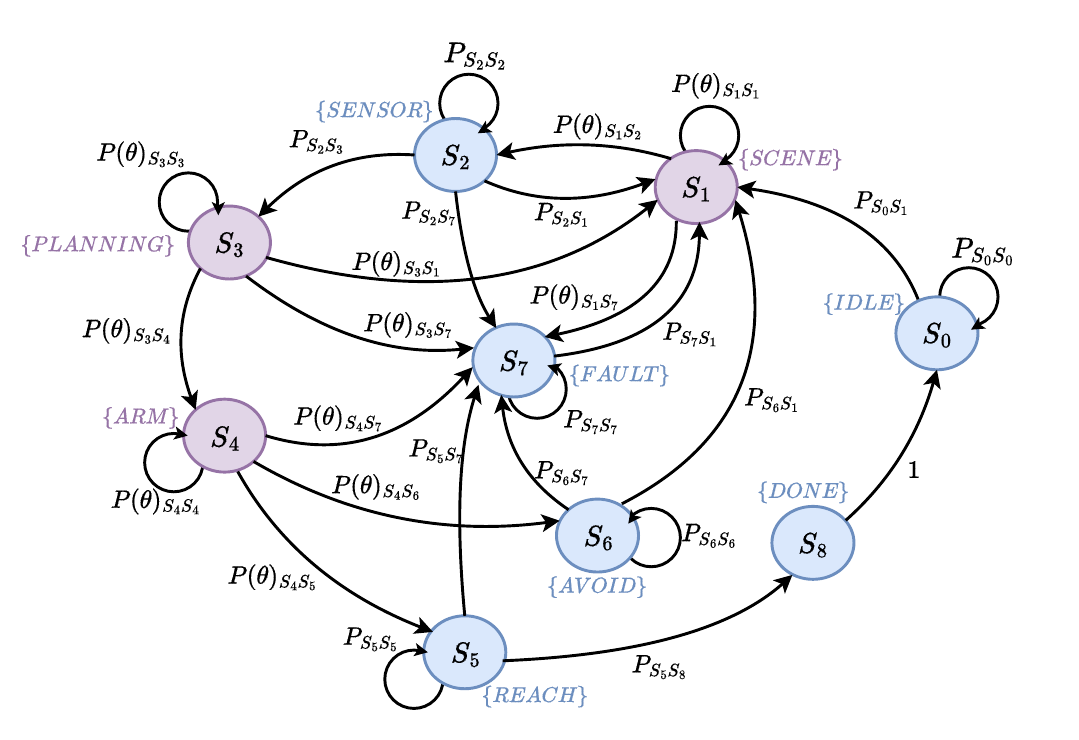}  
  \caption{System level parametric DTMC model: Blue colored states have constant outbound transition whereas purple colored states have parametric outbound transition probabilities.}
  \label{fig:sys_dtmc_big}
\end{figure*}

In our specific case, the system model, \(\mathcal{A}\), can be represented by a DTMC by adding probabilities to the transitions between states as illustrated in Figure~\ref{fig:sys_dtmc_big}.

We start by defining the states, $S$, and the labeling function, $L$. The state space, $S$ =\{\( S_0 \), \( S_1 \), \( S_2 \), \( S_3 \), \( S_4 \), \( S_5 \), \( S_6 \), \( S_7 \), \( S_8 \)\}, as well as the labeling function,  $L$, with atomic propositions set $AP=$\{IDLE, SCENE, SENSOR, PLANNING, ARM, REACH, AVOID, FAULT, DONE\}, are listed in Table~\ref{table:dtmc_states}. In our case, the initial state \( s_{\text{init}} = S_0  \), and the probability matrix, \(\mathbf{P} \),  will be parametric and therefore  denoted as \( \mathbf{P}_\theta \) in subsequent steps so that the probabilities of transition from one state to the other will depend on the design variant. The exact values of some of these probabilities are not entirely known at early design stage, but may need to be chosen based on data sheets, earlier or similar experiences, or crudely guesstimated. These assumed values are however to be verified or updated during the upcoming simulation and (indoor/outdoor) lab- and field testing activities normally occurring during the implementation and deployment phases of RAS engineering.

\begin{table}[htbp!]
\caption{System generic DTMC states, $S$, and corresponding labels, $L$.}
\centering
\label{table:dtmc_states}

 \begin{tabular}{|>\centering m{0.018\textwidth}| m{0.27\textwidth}|c |}  
  \hline
  $S$   & \multicolumn{1}{c|}{Description}          & Label, $L$  \\  
  \hline
  \hline
  \( S_0 \) & Initialization \& Idle, \( s_{\text{init}} = S_0 \)  & IDLE \\
  \hline
  \( S_1 \) & Scene perception & SCENE \\
    \hline
  \( S_2 \) & Processing sensor data  & SENSOR \\
    \hline
  \( S_3 \) & Motion planning & PLANNING \\
    \hline
  \( S_4 \) & Robot-arm controller &  ARM \\
    \hline
  \( S_5 \) & Reaching interaction point & REACH \\
    \hline
  \( S_6 \) & Avoiding obstacles  & AVOID \\
    \hline
  \( S_7 \) & Fault handling & FAULT \\
    \hline
  \( S_8 \) & Task  completion  & DONE \\
  \hline
\end{tabular}
\end{table}

\subsubsection*{System Variant Configuration}
 
According to our use case, system designers proposed design alternatives for certain sub-systems with varying performance parameters.
At this stage, we build the Design Space Tree, denoted \( \mathcal{DS} \), to capture all possible combinations of design variants. For the sake of simplicity, we assume that the choice of a design variant is independent from the rest of variant choices and that  all design variants are compatible with each other. This means that the size of \( \mathcal{DS} \) will be the product of sub-system variant alternatives.
We consider the following alternatives to address the scene perception, motion planning and manipulation problems.
\begin{itemize}
     \item Three potential solutions could be considered to solve the scene perception problem:
    \begin{itemize} 
       \item \(SP_1\): Single RGB camera sensor.
       \item \(SP_2\): Stereo (3D) camera sensors.
       \item \(SP_3\): Multi-modal sensing (RGB + point cloud).
    \end{itemize}
    \item Two alternatives for the motion planning component: 
    \begin{itemize} 
       \item \(MP_1\): Visual servoing algorithm 
       \item\(MP_2\): Multi-objective optimal control algorithm based on potential functions 
    \end{itemize}  
\item Two alternatives for Object Manipulation:   
    \begin{itemize}    
     \item \(OM_1\): One 6DOF robotic arm
     \item \(OM_2\): At least two 6DOF robotic arms
    \end{itemize}
\end{itemize}

\begin{figure*}[!htbp]
  \vspace{-0.2cm}
  \centering
   \includegraphics[width = 10cm]
   {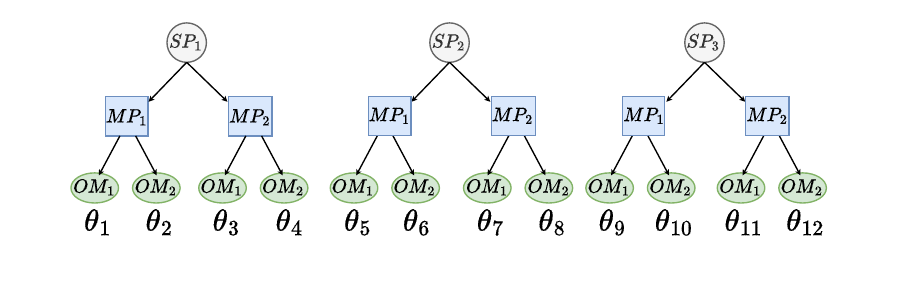}  
  \caption{System Design Configuration Tree, each branch in the graph, \( \theta_i \), represents a design variant from a combination of sub-system alternatives. }  
  \label{fig:sys_conf_tree}
\end{figure*}

Figure~\ref{fig:sys_conf_tree} illustrates how a typical combinatorial design tree is generated. Let \(n_{sc}\) denote the number of design alternatives of the scene perception sub-system, \(n_{mp}\) the number of design alternatives of the motion planning sub-system and \(n_{si}\) the number of design alternatives of the subject interaction sub-system. The size of the design space, \( \mathcal{DS} \), will then be \(N = n_{sc} \times n_{mp} \times n_{si} = 12 \). \( \mathcal{DS} \) will further contain the set of tuples representing the design alternatives, \( \mathcal{DS} = (\{ SP_i,MP_j,OM_k\})    , \forall i \in {1,2}, \text{and} j \in {1,2,3}.   \) 

Each element in $\mathcal{DS}$ thus represents a system variant configuration and  consequently induce a change in transition probability from the state affected by that sub-system to the rest of the states as demonstrated in Table~\ref{table:alternatives_vs_probabilities}. 

\begin{table}[htbp!]
\caption{Transition probabilities according to design alternatives}
\centering
\label{table:alternatives_vs_probabilities}

     \begin{tabular}{| m{0.06\textwidth}| m{0.06\textwidth}|m{0.06\textwidth}|  m{0.06\textwidth}|m{0.06\textwidth} |} %
      \hline
      \centering
       \( S_1 \)    & \(P_{S_1S_1} \)   & \(P_{S_1S_2} \)   & \(P_{S_1S_7} \)   &   \\ %
      \hline
      \(SP_1\)      & 0.496             & 0.496             &    0.008          &   \\
      \hline
      \(SP_2\)      & 0.2997            & 0.6993            &    0.001          &   \\
        \hline
      \(SP_3\)      & 0.04975           & 0.94525           &    0.005          &   \\
      \hline
        \end{tabular}
         
     \begin{tabular}{| m{0.06\textwidth}| m{0.06\textwidth}|m{0.06\textwidth}|  m{0.06\textwidth}|m{0.06\textwidth} |} %
      \hline
      \centering
      \( S_3 \)  &  \(P_{S_3S_3} \) & \(P_{S_3S_1} \)& \(P_{S_3S_4} \)  & \(P_{S_3S_7} \)   \\ %
      \hline
      \(MP_1\)   & 0.197            & 0.197          & 0.591             &    0.015          \\
      \hline
      \(MP_2\)  & 0.0495            & 0.099          & 0.8415           &     0.01          \\
        \hline
    \end{tabular}
    
     \begin{tabular}{| m{0.06\textwidth}| m{0.06\textwidth}|m{0.06\textwidth}|  m{0.06\textwidth}|m{0.06\textwidth} |} %
      \hline
      \centering
      \( S_4 \)   &  \(P_{S_4S_4} \)  & \(P_{S_4S_5} \) & \(P_{S_4S_6} \)  & \(P_{S_4S_7} \) \\ %
      \hline
      \(OM_1\)    & 0.2              & 0.3               &    0.3             &    0.2   \\
      \hline
      \(OM_2\)    & 0.04            & 0.45               &    0.45            &    0.06   \\
       \hline
    \end{tabular}

\end{table}

\subsection{Formal Specification and Modeling~\Circled[inner color=blue]{c}}
In this step, we perform the necessary transformation operations to achieve a formal representation of the system and the properties reflecting the verifiable criteria. Consequently we produce the necessary input artifacts that can be verifies by a probabilistic model checker, e.g., PRISM.

\subsubsection*{Design Criteria to Formal Properties Transformation }
\begin{table*}[htbp!]
\centering
\caption{PCTL property representation in PRISM syntax}
\label{table:property_prism}

 \begin{tabular}{|>\centering m{0.05\textwidth}| m{0.4\textwidth}|m{0.4\textwidth} |} %
  \hline
  \textbf{\( \Phi\)}   & \multicolumn{1}{c|}{\textbf{PCTL property description}}          & \textbf{PCTL property PRISM syntax}  \\ %
  \hline
  \( \varphi_1 \) &  What is the probability of reaching \{DONE\} without ever visiting \{FAULT\}?   & \begin{verbatim}P=?[(!s=FAULT)U s=DONE]\end{verbatim}  \\ 
  \hline
  \( \varphi_2 \) & What is the minimum expected number of time steps (reward) to reach state \{DONE\}?              & \begin{verbatim}R{"steps"}min=?[F s=DONE] \end{verbatim}\\ 
    \hline
  \( \varphi_3 \) & What fraction of time, in the long run, does the system spend in the \{DONE\} state?  & \begin{verbatim}S=? [ s=DONE ] \end{verbatim} \\ 
    \hline
  \( \varphi_4 \) & What fraction of time, in the long run, does the system spend in the \{FAULT\} state? & \begin{verbatim}S=? [ s=FAULT ] \end{verbatim} \\  
    \hline
  \( \varphi_5 \) & What fraction of time, in the long run, does the system spend in the \{FAULT\} state, directly coming from \{AVOID\} state? & \texttt{filter(state, P=? [X s=FAULT], s=AVOID)*R{"AVOID"}=?[S] } \\  
    \hline
    \( \varphi_6 \) & What fraction of time, in the long run, does the system spend in the \{FAULT\} state, directly coming from \{ARM\} state? & \texttt{filter(state, P=? [X s=FAULT], s=AVOID)*R{"ARM"}=?[S] } \\  
    \hline
   
  \hline
      \( \varphi_7 \) & What fraction of time, in the long run, does the system spend in \{SCENE\}? & \begin{verbatim}R{"scene_time"} =? [ S ]\end{verbatim} \\  
  \hline

\end{tabular}
\end{table*}

\begin{table}[htbp!]
\caption{Transformation from Design Criteria to Formal Properties}
\centering
\label{table:criteria_to_property}

 \begin{tabular}{|>\centering m{0.05\textwidth}| m{0.15\textwidth}|c |} %
  \hline
   \( \mathcal{VC} \)  & \multicolumn{1}{c|}{ \( \mathcal{F}_{\Phi} \) :\( \mathcal{VC} \) \(\rightarrow \) \( \Phi\)}           \\ %
  \hline
  \hline
  \(C_1\) & \( \varphi_1 \), \( \varphi_3 \), \( \varphi_4 \),\( \varphi_5 \), \( \varphi_6 \)     \\
  \hline
  \(C_2\) & \( \varphi_5 \)                \\
    \hline
  \(C_3\) & \( \varphi_7 \) \\  %
    \hline
  \(C_4\) &\( \varphi_2 \), \( \varphi_3 \), \( \varphi_4 \)  \\
    \hline
\end{tabular}
\end{table}

Table~\ref{table:property_prism} illustrates the property formulae set \( \Phi\) described in natural language and expressed in PRISM PCTL syntax. These properties were composed such that their respective verification results reflect one or multiple aspects of the Verifiable Criteria set, \( \mathcal{VC} \). In Table~\ref{table:criteria_to_property}, we list the properties that correspond to each criterion. The respective transformation functions, \( \mathcal{F}_{\Phi} \), can be as simple as identity or any form of weighted quadratic sum. It is noteworthy that the verification results can come in various data types, so that \( \mathcal{F}_{\Phi} \) must accommodate data conversion.
We also need to specify the Success Criteria set, denoted \( \mathcal{SC} \), which contains the threshold values that will be used to compare the property results against in order to decide whether to keep or discard a design variant after performing model checking. We chose to specify the following success thresholds: 
\begin{equation*}
     \mathcal{SC} = \{ \{ \varphi_1  \ge40\% \}, \{ \varphi_4  \le10\% \},  \{ \varphi_7  \le30\% \} \}.
\end{equation*}
We do not need to set a success threshold for every property verification result as the ranking step performed later in the process will eliminate the non-satisfactory design variants. Nevertheless, setting thresholds especially if they come as explicit design requirements would help keep design variant sets in a manageable size, especially if human assessment is needed for the evaluation and ranking process.

    \subsubsection*{Formal System Model} 
    We start from the established base DTMC system model, $\D$, to create a parametric PRISM script describing the system states and set all transitionary probabilities to variables. This step comes in preparation for the model checking step. PRISM~\cite{kwiatkowska2011} will be used as  back-end engine so that all transition properties will be passed as parameters.    
    In order to verify certain types of properties, namely those requiring calculating "steady-state" of the system over time, we need to add certain "reward" functions to trigger at certain events and allow PRISM to accumulate counter variables to emulate, for example, time elapsed starting from certain event or counting a number of activations of certain state.

\subsection{Probabilistic Model Checking~\Circled[inner color=blue]{d}}

The core implementation of the model checking phase is split into two main aspects: the PRISM model checker back-end and script written in python in our case that controls the PRISM via Command Line Interface instruction to automate the process. 
As the specific implementation of this mechanism can vary greatly, we briefly present the structure of our Python script implementing Algorithm~\ref{alg:mc_alg}. 
First, we instantiate the Execution wrapper, \( \mathcal{E} \), with reference to the parametric DTMC model file, \(D_{\theta} \), and the Formal property formulae set file, \( \Phi\). The execution wrapper, \( \mathcal{E} \), also sets all necessary links to the PRISM executable program depending on the operating system and its location in the file system.
Then we crate a dictionary of the \(12\) transition matrices corresponding the design alternatives of \( \mathcal{DS}\). The Python script runs PRISM verifications via the execution wrapper iteratively for each design alternative, \( \theta\).  At the end of each execution, we check the results or property, comparing, \(\mathcal{R} \), against the specified success criteria, \(\mathcal{SC}\) to determine if they belong to $\mathcal{R}_{success}$ or $\mathcal{R}_{fail}$. This constitutes a first layer of filtering in order to discard all infeasible design alternatives, failing to meet \(\mathcal{SC}\). In our practical experiment, the reduced sub-set retained from $\mathcal{R}_{success}$, include  the following variants:  
\begin{equation*}
\{ \theta_6, \theta_7, \theta_8, \theta_9, \theta_{10}, \theta_{12} \}.
\end{equation*}

\subsection{Reverse Specification Mapping~\Circled[inner color=blue]{e}}
The main purpose of this step it to map the values in \(\mathcal{R}_{succss} \) to a score system,  \( \mathcal{VC}_{score} \), that can be used for the ranking of the Verifies Design variants. There is no unique way of performing the mapping since it is dependent on multiple factors such as the types of property formulation, the data type of calculated results and the nature of the corresponding criteria. In this paper however, we adopt a series of post processing operations consisting of: \textit{value range mapping, normalization, scaling} to produce an appropriate score from the generated tabular artifact format of \(\mathcal{R}_{success} \). 

Value range mapping consist of adapting the value of the verification results to be within a bounded range compatible with the normalization step. The mapping also rectifies the value such that maximum value of the range represents the best performance, whilst the lowest represents the worst. This addresses, inter alia, the case where a property quantifies the degradation of the system. Properties \( \varphi_2\), \( \varphi_4\), \( \varphi_5\), \( \varphi_6\) and \( \varphi_7\) serve as typical examples, where a higher value mean worse performance. 
For percentage results, \(r_p\), which are inherently bound in the range \( [0,1] \), the complement function is sufficient to calculate the complement, \(r_{map} = (1- r_p)\).

Moving on to the normalization operations, as shown in Table~\ref{table:criteria_to_property}, criterion \(C_1\) depends on properties \( \{\varphi_1,\varphi_3,\varphi_4, \varphi_6\}\) which have unit-less percentages as their corresponding verification results. In general, the properties and their results may have different impact on the criterion they influence. Hence, a weighted sum serves as an appropriate approach to represent the effect of each property result. For this purpose, we define as set of weights, \(\alpha_i\), where:

\begin{equation*}
\sum_{i=1}^{n} \alpha_i = 1, \quad \alpha_i \geq 0.
\end{equation*}

Letting \( r_i \) denote the value of the calculated property, \( \varphi_i \), form the verification result as coming from the MC or after values remapping,  an intermediate weighted normalized score, \(s^\prime _1\),  is defines as:    
\begin{equation*}
s^\prime _1 = \sum_{i=1}^{n} \alpha_i r_i.
\end{equation*}
Since we want all criteria to have the same scale range, \( [0,10] \), we assign the value \(0\) to the worst performance representing the lowest respective percentage among the the range of design alternatives and \( 10 \) to the best. This is also appropriate since we are interested in the relative performance of the design variants rather than the absolute value. We apply a scaling function to \(s^\prime _1\) in order to calculate the scaled score,
\begin{equation*}
    s_1 = 10   \frac{s^\prime _1 - s_{1\min}} {s_{1\max} - s_{1\min}},
\end{equation*}
where:
\begin{itemize}
    \item \( s_{1\min} = \min_{\theta \in \Theta} s^\prime _1(\theta)  \),
     \item \( s_{1\max} = \max_{\theta \in \Theta} s^\prime _1(\theta)  \).
\end{itemize}

Regarding criterion \(C_2\) and \(C_3\), since each depend on only one property that returns a percentage of the verification result, we may just apply a scaling operation to transform the percentage into a score range of $[1, 10]$, consistent with the scaling operation performed on \(C_1\).
For \(C_4\), we apply the same post processing operations to the verification results of properties \( \varphi_2\), \( \varphi_3\) and \( \varphi_4\) to calculate the scores. 
At this stage, all scaled scores corresponding to each variant lie in the interval \([0 \;10]\) and are shown in Table~\ref{table:scaled_creteria}.

\begin{table}[htbp!]
\caption{Values of criteria \(C_1, \cdots, C_4\) and their average across design variants, \(\theta \in \Theta\).}
\centering
\label{table:scaled_creteria}
\begin{tabular}{|>{\centering\arraybackslash}m{0.05\textwidth}|>{\centering\arraybackslash}m{0.04\textwidth}|>{\centering\arraybackslash}m{0.04\textwidth}|>{\centering\arraybackslash}m{0.04\textwidth}|>{\centering\arraybackslash}m{0.04\textwidth}|>{\centering\arraybackslash}m{0.06\textwidth}|}
  \hline
  \(\theta\) & \(C_1\) & \(C_2\) & \(C_3\) & \(C_4\) & Avg.\ \\
  \hline
  \hline
  $\theta_6$  & 9.5 & 5.5 & 4.9 & 8.1 & 6.99 \\
  \hline
  $\theta_7$  & 1.1 & 3.9 & 3.9 & 3.3 & 3.03 \\
  \hline
  $\theta_8$  & 10.0 & 8.4 & 4.1 & 9.8 & 8.08 \\
  \hline
  $\theta_9$  & 0.3 & 2.7 & 0.7 & 2.1 & 1.42 \\
  \hline
  $\theta_{10}$ & 8.6 & 6.9 & 0.9 & 8.5 & 6.21 \\
  \hline
  $\theta_{12}$ & 9.1 & 10.0 & 0.2 & 10.0 & 7.34 \\
  \hline
\end{tabular}
\end{table}

\subsection{Ranking of Verified Design concepts~\Circled[inner color=blue]{f}} 

 As a last step in the methodology, we apply a basic MCDM process to select the most suitable Verified Design, \( \mathcal{VD}\), which generally can be one or multiple design alternatives selected according to the goals outlined in the design space phase.

\begin{figure}[!h]
  \vspace{-0.2cm}
  \centering
   \includegraphics[width = 6cm]{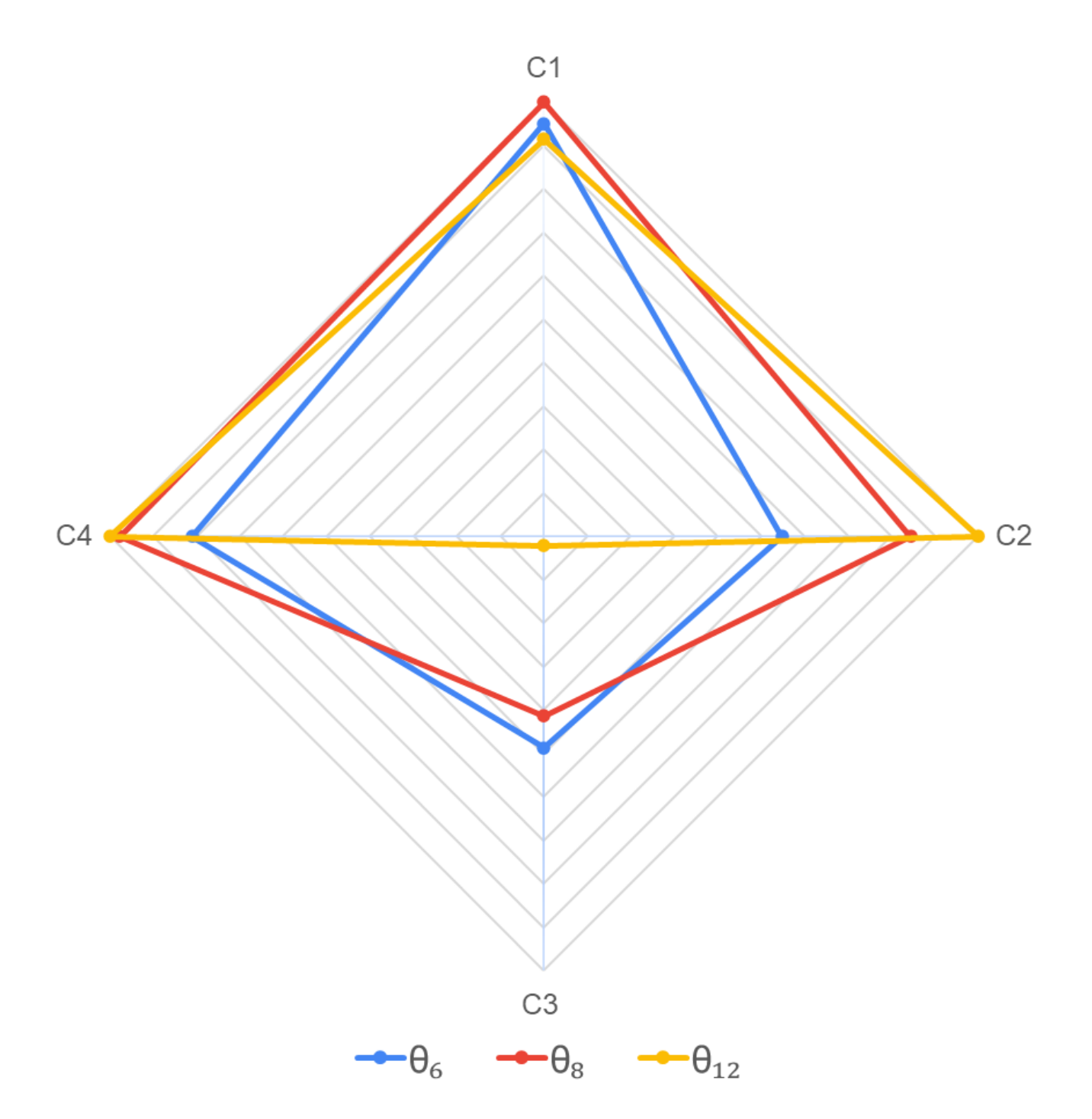}  
  \caption{Radar graph visualizing the scores of of the highest average ranking design variants respectively \(\theta_{6} , \theta_{8}, \theta_{12}\).}
  \label{fig:radar_graph}
\end{figure}

 In this paper, we rank the design variants only based on the formally verifiable criteria sub-set, \( \mathcal{VC}=\{C_1, C_2, C_3, C_4 \} \), in order to highlight the outcome of the PMC procedure. We consider a simple method of computing the average of the Verified Designs sub-set scores, \( \mathcal{VC}_{score} \), given that they are homogeneous and have the same scale range, and then select the top three highest averages. Design alternatives  
 \begin{itemize}
     \item \( \theta_{6} = \{SP2, \quad MP1, \quad OM1\} \),
     \item \( \theta_{8} = \{SP2, \quad MP2, \quad OM2\} \),
     \item \( \theta_{12} = \{SP3, \quad MP2, \quad OM2\} \),

 \end{itemize}
 have the three highest score averages across \( \mathcal{VC} \).
 However, in a realistic scenario, \( \{ C_5...C_9 \} \) have to be considered as well, so that the ranking and selection of design variants take all aspects into consideration.   
 At this stage, more advanced and elaborate visual tools, such as radar graphs as illustrated in Figure~\ref{fig:radar_graph} can be used to display the design alternatives scores, so that stakeholders have more intuitive overview of the strengths and weaknesses of each Verified Design variant in \( \mathcal{VD} \) to allow for more informed decisions.

By applying the average of criteria scores, design variant \( \theta_{8} \) ranked the highest, however \( \theta_{12} \) the highest in \(  C_2 \) and \(  C_4 \) but had the lowest score in \( C_3 \). This imbalance could signify a specific anomaly in that design variant and calls for further investigation by the design team. Eventually some adjustments could be performed to mitigate short comings of certain design variant, then running PMC again in order to make sure the adjusted design remain verified.

\section{\uppercase{Concluding remarks and future work}}
\label{sec:conclusion}

Recognizing the implications and effect of system design and concept selection on safety and reliability aspects of RAS, this paper proposes a novel approach to add a formal verification layer to the design space exploration step in the concept study phase. We start by generating the initial design alternatives, then systematically run probabilistic model checking on every variant and finally arrive at the set of Verified Design (VD). 

We applied our methodology to a use case from agriculture robotics where system behavior was modeled by a DTMC, although other types of Markovian representations (e.g., Markov Decision Processes (MDP)) can be used, depending on the need and the specifics of the system.

The methodology can also be generalized to automatically translate SysML activity diagrams to Markov chain representations.  
Another future consideration is to develop a software tool to automate all steps described in the proposed methodology to achieve a seamless workflow from the process model (P2) stage resulting in a Verified Design (P3) supported by necessary artifacts proving its correctness.

\section*{\uppercase{Acknowledgments}}
 
This work has received partial funding from the Norwegian Research Council (RCN) RoboFarmer project number 336712

\bibliographystyle{apalike}
{\small
\bibliography{bibliography}}

\end{document}